\documentclass[titlepage,11pt]{article}
\usepackage{PRIMEarxiv}
\usepackage[utf8]{inputenc} 
\usepackage[T1]{fontenc}    
\usepackage{hyperref}       
\usepackage{url}            
\usepackage{booktabs}       
\usepackage{amsfonts}       
\usepackage{nicefrac}       
\usepackage{microtype}      
\usepackage{lipsum}
\usepackage{fancyhdr}       
\usepackage{graphicx}       
\graphicspath{{media/}}     
\usepackage{indentfirst}
\usepackage{setspace}
\begin{document}
\begin{flushleft}
\centerline{\Large \textbf \newline{BHGNN-RT: Network embedding for directed heterogeneous graphs}} 
\bigskip
\centerline{Xiyang Sun\textsuperscript{1}, Fumiyasu Komaki\textsuperscript{1,2*} }
\textbf{1}. Mathematical Informatics Collaboration Unit, 
RIKEN Center for Brain Science, Japan. \\
\textbf{2}. Department of Mathematical Informatics, 
The University of Tokyo, Japan. \\
* Corresponding author: komaki@g.ecc.u-tokyo.ac.jp 
\end{flushleft}

\begin{abstract}
\indent Networks are one of the most valuable data structures for modeling problems in the real world. However, the most recent node embedding strategies have focused on undirected graphs, with limited attention to directed graphs, especially directed heterogeneous graphs. In this study, we first investigated the network properties of directed heterogeneous graphs. Based on network analysis, we proposed an embedding method, a bidirectional heterogeneous graph neural network with random teleport (BHGNN-RT), for directed heterogeneous graphs, that leverages bidirectional message-passing process and network heterogeneity. With the optimization of teleport proportion, BHGNN-RT is beneficial to overcome the over-smoothing problem. Extensive experiments on various datasets were conducted to verify the efficacy and efficiency of BHGNN-RT. Furthermore, we investigated the effects of message components, model layer, and teleport proportion on model performance. The performance comparison with all other baselines illustrates that BHGNN-RT achieves state-of-the-art performance, outperforming the benchmark methods in both node classification and unsupervised clustering tasks. 
\end{abstract}

\section{Introduction}
\indent Many real-world systems come with graph forms, such as citation networks, social networks, and the World Wide Web \cite{wang2019heterogeneous, zhao2021heterogeneous}. Graphs possess intrinsic and strong capabilities to represent complex structures and can easily express entities and their relationships \cite{tong2020directed}. It is widely recognized that graph data are often sophisticated and, thus, challenging to process \cite{cui2018survey}. A graph neural network (GNN) is a powerful graph representation learning method designed for such graph data and has attracted considerable research attention \cite{zhang2019graph, mavromatis2023global}. Traditionally, GNNs focus on individual nodes to generate a vector representation or an embedding for each node, such that two nodes “close” in the graph have similar vector representations in a low-dimensional space \cite{zheng2016node}. Recently, many variants of GNNs have achieved superior performances in network analysis, including node classification \cite{hamilton2017inductive}, graph classification, link prediction \cite{schlichtkrull2018modeling}, and recommendations. Examples include spectral graph convolutional neural networks \cite{defferrard2016convolutional, kipf2016semi, wu2019simplifying}, message-passing algorithms \cite{velickovic2018graph} and recurrent graph neural networks \cite{ioannidis2019recurrent}. Among them, message-passing frameworks have received particular attention because of their flexibility and good performance \cite{kipf2016semi, gasteiger2018predict}. \\
\indent Among the information encoded in a graph, graph structures and properties primarily affect graph inference \cite{cui2018survey}. Hence, preserving these two factors is necessary for graph representation learning. However, existing GNNs, particularly spectral-based GNNs, mainly consider undirected graph scenarios \cite{schlichtkrull2018modeling, busbridge2019relational, tong2020directed}. As a matter of fact, most real-world graphs are directed. For example, in a citation network, more recent papers can cite older ones; however, the opposite is not true. These incoming and outgoing connections can express completely different relationships and meanings in a directed graph. Recently, some studies proposed GNN methods for directed graphs, such as DGCN \cite{tong2020directed}, NERD \cite{khosla2020node}, MPAD \cite{nikolentzos2020message}, and Dir-GCN \cite{rossi2023edge}. These GNNs leverage incoming messages and second-order proximity or adopt input neighborhood sampling while ignoring these edge direction information and asymmetry between incoming and outgoing connections. Integrating this bidirectional information shall provide a more comprehensive representation of the local structure around a node, which is particularly important in directed graphs because they can indicate different meanings and functional roles. \\
\indent Another common issue in GNNs is over-smoothing. Theoretically, the message-passing process of $k$ iterations takes advantage of a subtree structure with height $k$ rooted at each node. Such schemes can generalize the Weisfeiler-Lehman graph isomorphism test to learn the distribution and topology of node features in the neighborhood simultaneously \cite{hamilton2017inductive, xu2018representation}. However, increasing the number of iterations with too many layers usually leads to over-smoothing. For example, previous work indicated that the best performance of a SOTA model, the graph convolutional network (GCN), is achieved with a 2-layer structure \cite{kipf2016semi}. Their embedding results converged to the random walk's limit distribution as the layer number increased \cite{kipf2016semi, gasteiger2018predict}. Other methods have also been faced with the same problem \cite{xu2018representation, klicpera2019combining, rong2019truly}. In principle, deeper versions of GCN perform worse, although they have access to more information. The limitation of GNN layer configuration strongly restrains the expressivity of node neighborhoods with high path lengths. To solve this issue, we introduce the random teleport into our model and optimize the teleport proportion while updating node embedding with bidirectional information. The teleport proportion helps to counteract this convergence by resetting a random initial state when we preserve the node locality \cite{gasteiger2018predict, roth2022transforming}.\\
\indent In this work, we first characterized the network properties of directed heterogeneous graphs, including asymmetric degree distribution and network heterogeneity. The appropriate preservation of these factors allows better graph representation learning. Motivated by network analysis, we proposed a novel GNN model called a bidirectional heterogeneous graph neural network with random teleport (BHGNN-RT) to leverage the bidirectional message-passing process and network heterogeneity. With random teleport, the message-passing process will not be easily trapped into nodes with self-loops or without outgoing edges in the directed graph. The optimization of teleport proportion allows balancing messages from existing neighborhoods with random connections, which is beneficial for overcoming over-smoothing in the node embedding. Afterward, we conducted extensive experiments on benchmark datasets to validate the effectiveness of BHGNN-RT compared with the benchmark algorithms. Experimental results show that BHGNN-RT obtains higher accuracies in different tasks and achieves state-of-the-art performance. We further investigated the effect of message components, model layer, and teleport proportion on the model performance. The experimental results demonstrate that BHGNN-RT allows more accurate modeling of directed heterogeneous graphs, covering a broad class of application scenarios. \\
\indent The rest of our paper is organized as follows. Section 2 discusses the problem formation for representing network features in directed heterogeneous graphs, which inspires our modeling work. Section 3 mainly presents our proposed model, BHGNN-RT, for network embeddings. We outline our experiments in Section 4 and display the corresponding experimental results in Section 5. We close with the conclusion of our work in Section 6.

\section{Problem Formation}
This section first introduces the concepts and preliminary knowledge of directed heterogeneous graphs. Subsequently, the relevant network properties were investigated to inspire our next work on network embedding. Through this paper, the matrices, vectors, and scalars are denoted by bold capital letters, bold lowercase letters, and lowercase letters, respectively.
\subsection{Directed heterogeneous graph}
Our target of interest is the directed heterogeneous graph $\mathcal{G} = \{\mathcal{V}, \mathcal{E}, \mathcal{T}, \mathcal{R}, \textbf{A} \}$, which consists of a set of nodes $\mathcal{V}$ with $|\mathcal{V}|=n$, a set of edges $\mathcal{E}$ with $|\mathcal{E}|=m$, a set of node types $\mathcal{T}$, a set of edge relations $\mathcal{R}$, and the adjacency matrix $\textbf{A}$. A heterogeneous graph contains either multiple types of nodes or edges; thus, $|\mathcal{T}|+|\mathcal{R}|>2$. If there is an edge from node $j$ to $i$, element $\textbf{A}_{ij}$ denotes the weight of this edge; otherwise, $\textbf{A}_{ij}=0$. For unweighted graphs, $\textbf{A}_{ij}$ is simply configured as 1. The node attributes or features are denoted as $\textbf{X} \in \mathbb{R}^{n \times f}$, where $\textbf{x}_i$ is the initial feature of node $i$ and its column number $f$ is the feature dimension.
\subsection{Network properties of directed heterogeneous graphs}
\indent When investigating the network properties, the most basic and intuitive measure is its degree or weight distribution. Here, we characterize the properties of directed heterogeneous graphs, mainly from degree distribution and network heterogeneity. To ensure the reliability and robustness of our study, data diversity is satisfied by selecting diverse benchmark datasets, namely Cora, Cora\_ml, CiteSeer, CiteSeer\_full, Amazon\_cs, and Amazon\_photo, with different sizes and various edge relationships. Detailed descriptions of these datasets are provided in the Supplementary Materials. Our analyses reflect the necessity to pay attention to these network properties in graph representation learning. 
\subsubsection{Asymmetry between in-degree and out-degree distributions}
\begin{figure}
    \centering
   \includegraphics[scale=0.48]{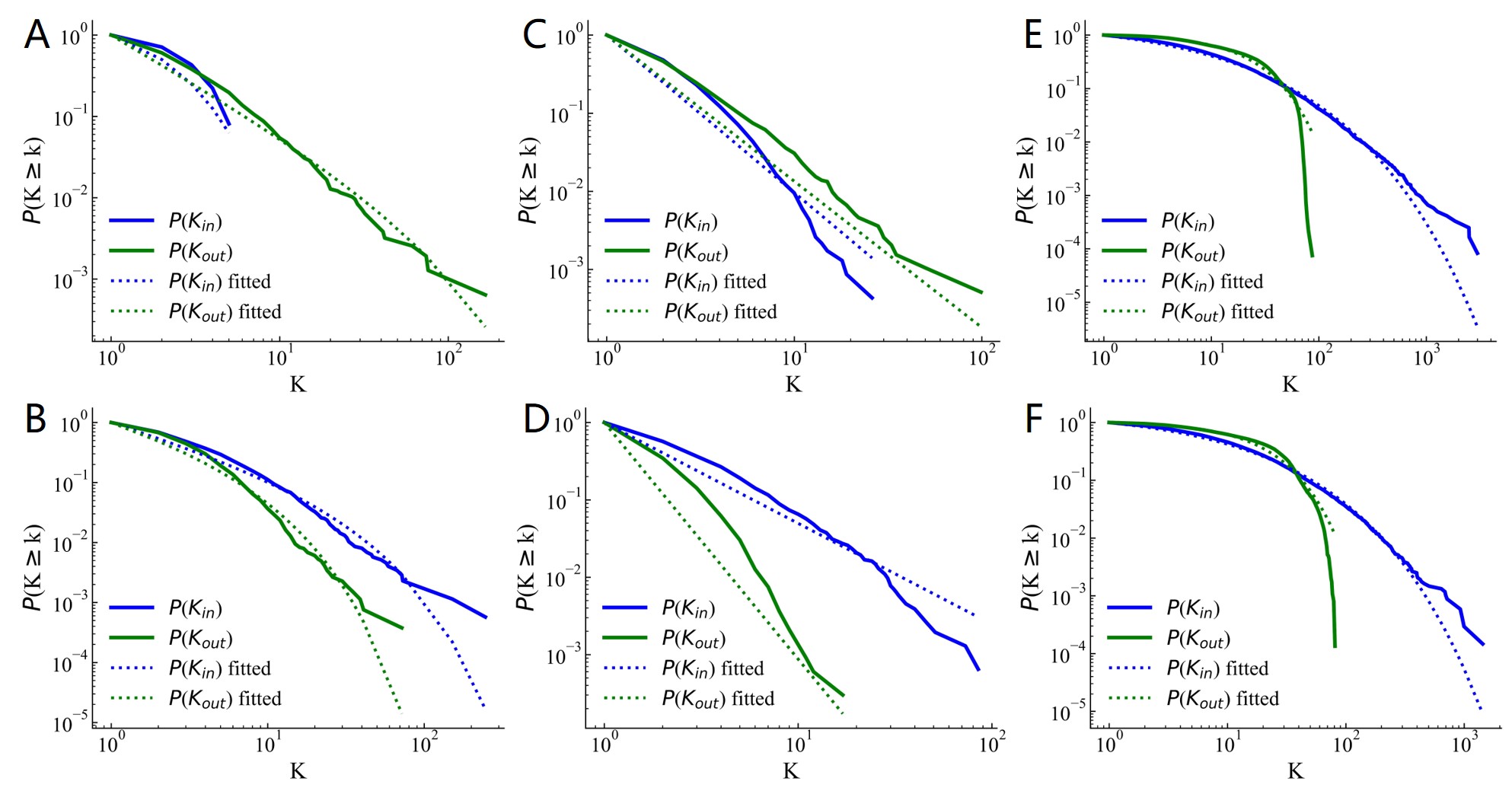}
    \caption{\textbf{Asymmetry between in- and out-degree distributions.} The datasets correspond to Cora (A), Cora\_ml (B), CiteSeer (C), CiteSeer\_full (D), Amazon\_cs (E), and Amazon\_photo (F). Corresponding fits of in- and out-degree distributions are depicted as dashed lines and explained in Table \ref{Table 1}}.
    \label{Figure 1}
\end{figure}

\indent Compared with undirected networks, the degree distribution of a directed network is more complicated since the degree of a node cannot be fully captured by one single number. In directed graphs, distinguishing between in-degree and out-degree distributions is important \cite{topirceanu2018weighted, steinbock2019analytical}. We denote the in- and out-degree distribution of a network as $P_{\mathrm{deg}}(K_{\mathrm{in}}=k)$ and $P_{\mathrm{deg}}(K_{\mathrm{out}}=k)$, respectively. The means of the in- and out-degree distributions are equal, suggesting that $\langle K_{\mathrm{in}} \rangle=\langle K_{\mathrm{out}} \rangle$, However, the in-degree and out-degree distributions usually display completely different patterns in real-world networks. \\
\indent Here, we depict the degree distributions with complementary cumulative distribution functions (CCDF), plotted on logarithmic axes. The overall appearance of these distributions and scaling behaviors are more apparent via CCDFs, which are less noisy than the degree distribution plots. Logarithmic axes help visualize heavy-tailed distributions. CCDF is defined as 
\begin{equation}
\label{eq:1}   F(k)=\sum_{K=k}^{\infty} P_{\mathrm{deg}}(K) 
\end{equation}
\noindent which sums the probability for a node with a degree larger than $k$. By definition, $F(0)=1$ and $F(k_{max}+1)=0$, and $F(k)$ decreases monotonically with $k$. The degree distributions of these directed heterogeneous graphs are displayed in Figure \ref{Figure 1}. We fit the in- and out-degree distributions with five common distribution functions via the Power-law package \cite{alstott2014powerlaw}. These distribution functions are described in the Supplementary Table \ref{Supplementary Table 2}.
In addition, Akaike's information criterion (AIC) is applied to determine the best fit among all distribution candidates, including power law, truncated power law, exponential, stretched exponential, and lognormal functions.
AIC is defined by
$\mathrm{AIC} = 2k -2LL$,
where $k$ is the number of estimated parameters and $LL$ is the log-likelihood score for the model \cite{anderson2004model}. Detailed fitting results are presented in Table \ref{Table 1}, where a smaller value indicates a better fit. \\
\indent In Figure \ref{Figure 1}, panels A-D represent citation networks, while panels E-F represent co-purchasing networks in Amazon. Only Cora\_ml and Amazon\_photo display consistent in- and out-degree distribution patterns. Although the optimal fit of the in-degree distribution of CiteSeer\_full is lognormal (Table \ref{Table 1}), we select the suboptimal fit as the power-law distribution because an underflow error occurs when applying the optimal fit. Hence, CiteSeer\_full also exhibits similar in- and out-degree distribution patterns. However, the other three datasets present asymmetry within the in- and out-degree distributions. The asymmetric pattern has also been observed in many other networks, such as gene regulatory and industrial networks \cite{luo2015asymmetry, ichinose2018asymmetry}. The variations within the in- and out-degree distributions illustrate that the incoming and outgoing edges capture different relationships and topological information \cite{luo2015asymmetry, huang2017monoamine}. Therefore, it is highly necessary to capture these asymmetric patterns during the network embedding for directed graphs.

\begin{table}[!ht]
\small
\centering
\caption{Model selection for degree distribution based on Akaike's information criterion}
\begin{tabular}{cc|ccccc|c}
\toprule  
               & & Power-law & Truncated & Exponential & Stretched   & Lognormal &  Fitting function\\
               & &           & power-law &             & exponential &           & \\
\midrule  
Cora           & $P(K_{\mathrm{in}})$  & 6409.0 & 6104.2 & \bf6078.6 & 6078.9 & 6238.3 & Exponential \\
               & $P(K_{\mathrm{out}})$ & 4860.1 & \bf4857.3 & 5962.9 & 4886.6 & 4862.1 & Truncated power-law \\
\midrule  
Cora\_ml       & $P(K_{\mathrm{in}})$  & 7148.3 & \bf7100.9 & 8176.3 & 7129.5 & 7135.5 & Truncated power-law \\
               & $P(K_{\mathrm{out}})$ & 8937.4 & \bf8804.3 & 9421.8 & 8854.6 & 8880.6 & Truncated power-law \\
\midrule  
CiteSeer       & $P(K_{\mathrm{in}})$  & 3679.4 & 3681.4 & 4804.7 & 3753.3 & \bf3677.4 & Lognormal \\
               & $P(K_{\mathrm{out}})$ & \bf3545.9 & 3547.9 & 5265.2 & 3639.8 & 3547.1 & Power-law \\
\midrule  
CiteSeer\_full & $P(K_{\mathrm{in}})$  & \bf4648.8 & 4649.0 & 5886.5 & 4698.0 & \bf4647.1(*) & Power-law\\
               & $P(K_{\mathrm{out}})$ & \bf1394.4 & 1396.4 & 3374.0 & 1606.9 & 1396.8 & Power-law \\
\midrule  
Amazon\_cs     & $P(K_{\mathrm{in}})$  & 92423.2 & 89813.2 & 99834.7 & \bf89640.1 & 89903.5 &  Stretched exponential\\
               & $P(K_{\mathrm{out}})$ & 121849.7 & 112562.7 & \bf107557.4 & 107558.4 & 109371.2 & Exponential \\
\midrule  
Amazon\_photo  & $P(K_{\mathrm{in}})$  & 52486.8 & 50545.7 & 54485.1 & \bf50490.2 & 50690.8 & Stretched exponential \\
               & $P(K_{\mathrm{out}})$ & 65836.3 & 60817.8 & 58214.9 & \bf58213.2 & 59429.3 & Stretched exponential \\
\bottomrule 
\end{tabular}
\label{Table 1}
\begin{flushleft}Table 1 depicts the AIC scores for fitting in- and out-degree distributions with different distribution candidates, including power-law, truncated power-law, exponential, stretched exponential, and lognormal functions. A smaller score indicates a better fit. The underflow error occurs in the bold-starred case, where we lack the numerical precision to measure the extreme results of the optimal fit. Therefore, we select the sub-optimal fits for the in-degree distribution of CiteSeer\_full. The final column is the fitting function for the corresponding degree distribution in Figure \ref{Figure 1}. 
\end{flushleft}
\end{table}

\subsubsection{Graph heterogeneity}
\begin{figure}
    \centering
    \includegraphics[scale=0.48]{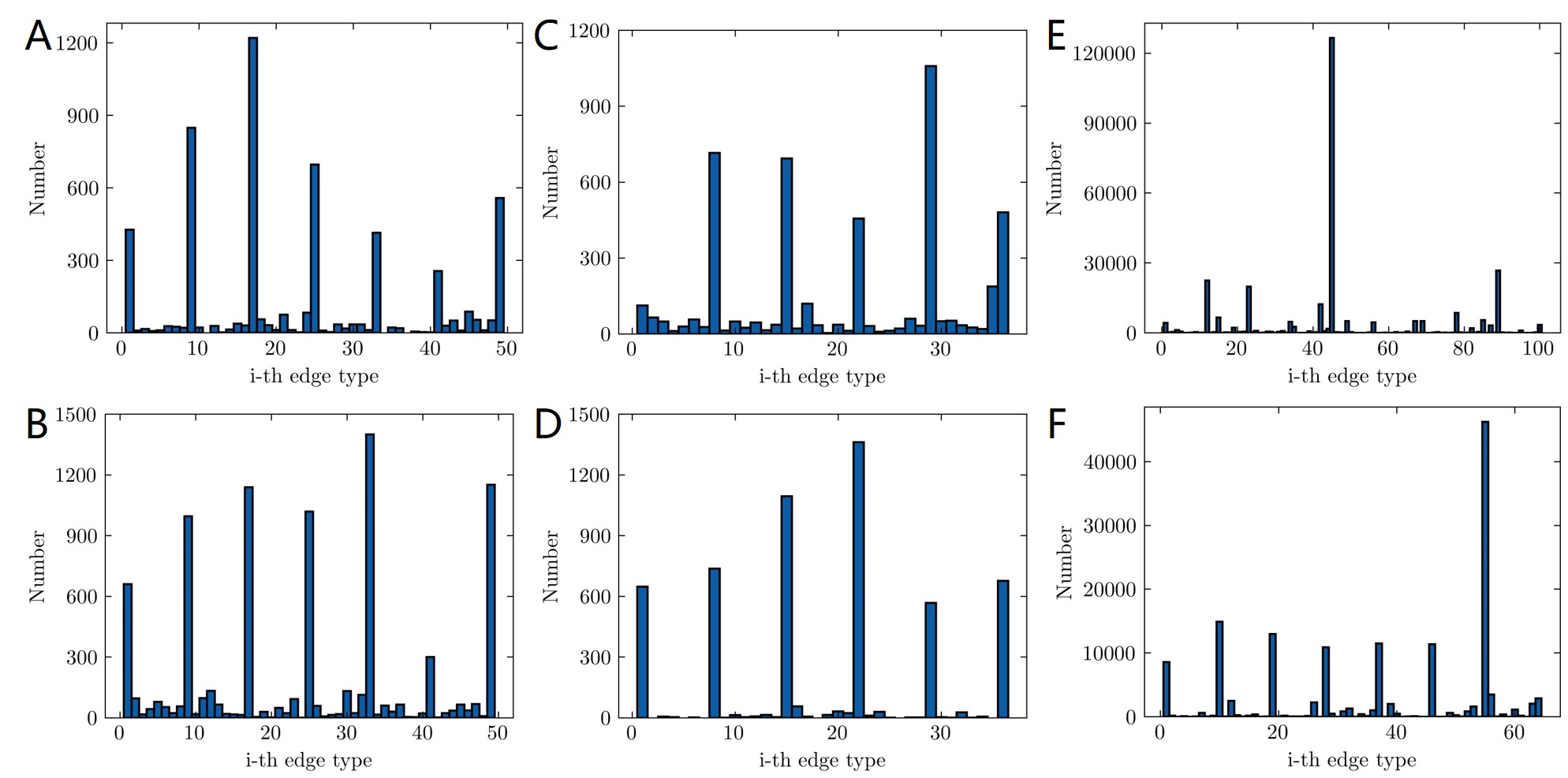}
    \caption{\textbf{Heterogeneity of edge relations.} The datasets correspond to Cora (A), Cora\_ml (D), CiteSeer (B), CiteSeer\_full (E), Amazon\_cs (C), and Amazon\_photo (F). The horizontal axis represents the $i$-th edge type in the graph.}
    \label{Figure 2}
\end{figure}

\indent An intrinsic property of the heterogeneous graph is heterogeneity \cite{wang2019heterogeneous}, referring to different kinds of nodes and edges with different attributes. Generally speaking, a heterogeneous graph has diverse edge relations, each reflecting an individual connection pattern. As illustrated in Figure \ref{Figure 2}, all benchmark datasets have multiple edge types. In each panel, the distribution of the edge types is imbalanced. The minority of the set of edge relations $\mathcal{R}$ usually occupies a large proportion of the whole edges, while the majority are only with small proportions. This indicates variations in the prevalence of edge relations, and prevalent edge types shall play a more important role in network connectivity. Similarly, the node type also possesses this kind of imbalanced heterogeneity, as shown in Supplementary Figure \ref{Supplementary Figure 1}. Hence, the performance of graph representation learning shall be strongly restricted, unless we pay attention to the graph heterogeneity. \\
\indent Although various graph representation learning methods have been proposed for adaptively learning graph structures and node features, most are designed for homogeneous graphs and cannot be directly applied to heterogeneous graphs \cite{zhao2021heterogeneous}. Dealing with such complex graph structures while preserving diverse information is still an urgent problem that must be solved \cite{wang2019heterogeneous}.

\section{Network embedding strategy}
Inspired by the network analyses, our network-embedding strategy is elaborated in this section for the directed heterogeneous graphs. Following relevant work on GNNs, we explain the details of the proposed method. Subsequently, we introduce objective functions for different experimental tasks to optimize the parameter spaces.  
\subsection{Graph Neural Network}
\indent Most GNNs operate under the message-passing framework \cite{schlichtkrull2018modeling, zhang2019graph}, where vector messages are exchanged between nodes and updated layer by layer. Given a graph $\mathcal{G} = \{\mathcal{V}, \mathcal{E} \}$ and node features $\textbf{X}\in \mathbb{R}^{n\times f}$, a GNN can use this information to generate the vector representations or embeddings $\textbf{z}_i$, $\forall i\in \mathcal{V}$. During each message passing iteration in the GNN, a hidden embedding $\textbf{h}_i^{l}$ is updated according to the messages aggregated from its neighborhood $N(i)$. Subsequently, the nodes can accumulate insights from their surroundings and capture local and global patterns \cite{kipf2016semi}. \\
\indent Typically, the message-passing process can be described as follows:
\begin{equation}
\label{eq:3}   \textbf{m}^{l+1}_j = \mathrm{MESSAGE}(\textbf{h}^l_j), j\in \{N(i) \cup i\} 
\end{equation}
\begin{equation}
\label{eq:4}   \textbf{h}^{l+1}_i = \sigma(\mathrm{AGGREGATE}(\{ \textbf{m}^{l+1}_j, j\in{N(i)} \}, \textbf{m}^{l+1}_i) ).
\end{equation}
Here, $\textbf{h}^l_i \in \mathbb{R}^{d^l}$ is the hidden feature of node $i$ in the $l$th layer of the model and $d^l$ is the dimensionality of the corresponding layer
where $l = 1, \ldots, L$ is the iteration number of the GNN layer, and
$N(i)$ defines a set of nodes adjacent to node $i$. $\mathrm{MESSAGE}(\cdot)$ and $\mathrm{AGGREGATE}(\cdot)$ are arbitrary differentiable functions. $\textbf{m}_j$ denotes the message aggregated from $i$'s neighborhood $N(i)$. $\mathrm{MESSAGE}(\cdot)$ refers to a message-specific neural network-like function or a simple linear transformation $\mathrm{MESSAGE}(\textbf{h}^l_j)=\textbf{m}^{l+1}_j=\textbf{W}^l\textbf{h}^l_j$, where $\textbf{W}^l$ is a trainable weight matrix on the $l$-th layer shared by all nodes \cite{zhang2019graph, kipf2016semi}. $\mathrm{AGGREGATE}(\cdot)$ represents the aggregator function, such as $\mathrm{Sum}(\cdot), \mathrm{Mean}(\cdot)$, or $\mathrm{Max}(\cdot)$. $\sigma(\cdot)$ is a nonlinear activation function (e.g., sigmoid function). The initial embeddings of the input layer are configured as node features, that is, $\textbf{h}_i^0=\textbf{x}_i$. After running $L$ iterations of message passing, the output of the final layer is generated as the final embedding for each node, that is, $\textbf{z}_i=\textbf{h}_i^L$. This framework has achieved significant performance in diverse tasks, such as graph classification \cite{kipf2016semi} and graph-based semisupervised learning \cite{schlichtkrull2018modeling, gasteiger2018predict}, node clustering \cite{mavromatis2020graph, zang2021unsupervised} and recommendation systems \cite{cui2018survey, zhang2019graph}.

\subsection{BHGNN-RT framework}
Inspired by our network analysis(Section 2.2), we proposed a new node-embedding algorithm, which captures the bidirectional message-passing process and network heterogeneity, for directed heterogeneous graphs.

\begin{figure}
    \centering
    \includegraphics[scale=0.5]{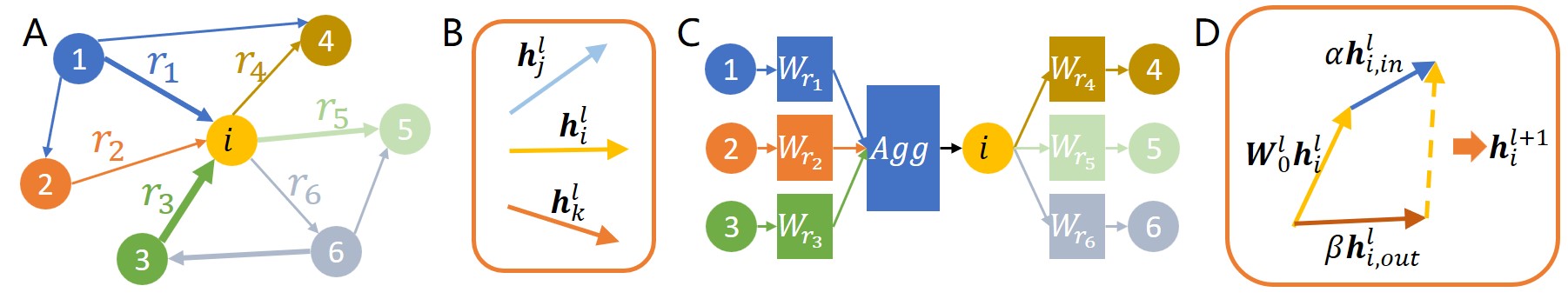}
    \caption{\textbf{Illustration of the node representation in BHGNN-RT framework.} Panel A depicts an example of a directed heterogeneous graph. B categorizes the neighborhoods ($j\in N_{in}(i), k\in N_{out}(i)$) of node $i$, according to the incoming and outgoing edges. C describes the main message-passing process for node $i$. D illustrates the updation function without the teleport component for node $i$.}
    \label{Figure 3}
\end{figure}

First, we categorize the edges attached to node $i$ into incoming and outgoing edges. The sets of nodes adjacent to node $i$ with incoming and outgoing edges are defined as $N_{\mathrm{in}}(i)$ and $N_{\mathrm{out}}(i)$, respectively (Figure \ref{Figure 3} B). An edge-type-dependent attention mechanism was introduced to handle network heterogeneity. The message function is adjusted using different weight matrices $\textbf{W}_r$ based on edge relation $r$. Meanwhile, the message depends on edge weight. Along the edge from a source node $j$ to a target node $i$ with weight $\textbf{A}_{ij}$, the overall input weight of node $i$ is $\textbf{A}_{i,\mathrm{in}}=\sum^N_{j=1} \textbf{A}_{ij}$ and the overall output weight of node $j$ is $\textbf{A}_{j,\mathrm{out}}=\sum^N_{k=1} \textbf{A}_{kj}$. To reduce message sensitivity to edge-weight scaling, the message from node $j$ to $i$ is normalized by the coefficient $\frac{\textbf{A}_{ij}}{\sqrt{\textbf{A}_{i, \mathrm{in}}} \sqrt{\textbf{A}_{j, \mathrm{out}}}}$. For an unweighted graph, the normalization coefficient is $\frac{1}{\sqrt{deg^-(i)} \sqrt{deg^+(j)}}$ \cite{kipf2016semi}, where $deg^-(\cdot)$ and $deg^+(\cdot)$ are the nodal in- and out-degrees, respectively. Then, we take the sum of the incoming messages under different edge relations as follows:  
\begin{equation}
\label{eq:5}
    \textbf{h}_{i,\mathrm{in}}^{l} = \sum_{r\in R}\sum_{j\in{N^r_{\mathrm{in}}(i)}} \frac{\textbf{A}_{ij}}{\sqrt{\textbf{A}_{i,\mathrm{in}}} \sqrt{\textbf{A}_{j,\mathrm{out}}}} \textbf{W}^{l}_{r} \textbf{h}^{l}_j 
\end{equation}
\noindent Regarding the outgoing message, the aggregation function is the weighted summation of the nodal hidden state $\textbf{h}^l_i$ instead of the hidden states of node $i$'s outgoing neighborhood. The outgoing messages from node $i$ are computed as: 
\begin{equation}
\label{eq:6} 
    \textbf{h}_{i,\mathrm{out}}^{l} = \sum_{r\in R}\sum_{k\in{N^r_{\mathrm{out}}(i)}} \frac{\textbf{A}_{ki}}{\sqrt{\textbf{A}_{k,\mathrm{in}}} \sqrt{\textbf{A}_{i,\mathrm{out}}}} \textbf{W}^{l}_{r} \textbf{h}^{l}_i 
\end{equation}

Afterward, each node aggregates the incoming and outgoing messages and the nodal message from itself. Considering the diverse roles of incoming and outgoing messages, we assigned them different weight coefficients that can be optimized during the learning process. The node embedding is then updated with a linear combination of all message components transformed in the $l$th layer.
\begin{eqnarray}
\label{eq:7}
    \textbf{h}_{i}^{l+1} = \sigma(\textbf{W}^l_0 \textbf{h}^l_i +\alpha \textbf{h}^l_{i,\mathrm{in}} -\beta \textbf{h}^l_{i,\mathrm{out}})
\end{eqnarray}
where the activation function $\sigma$ is a parametric rectified linear unit (PReLU), $\sigma(\cdot)=\mathrm{max}(0,\cdot)+ a \mathrm{min}(0,\cdot)$ with a learnable parameter $a$. The weights $\alpha$ and $\beta$ are message coefficients trained in the experimental tasks. The implementation of Eq \ref{eq:7} in the vectorized form requires computationally efficient sparse dense $\mathcal{O}(|\mathcal{E}|^2)$ matrix multiplications.

Besides, we do not expect the message-passing process shall be trapped in some nodes in the directed heterogeneous graph. Typically, nodes with strong self-loops or without outgoing edges easily absorb incoming messages and have little interaction with other nodes. In this case, the message-passing process does not converge to the embedding results we want. To overcome this problem, we draw inspiration from personalized PageRank \cite{page1998pagerank} and introduce a teleport vector $\overrightarrow{\textbf{1}} \in \mathbb{R}^{d^{l+1}}$ into the aggregator function. We assigned the probability $\gamma \in [0, 1]$ to the teleport proportion, allowing node $i$ to receive messages from random connections. With a probability $1-\gamma$, node $i$ obtains messages from existing connectivities. Moreover, unlike Gasteiger's work using a one-hot indicator vector for the teleport vector \cite{gasteiger2018predict}, we need not normalize the adjacency matrix or add self-loops. The aggregator function with random teleport is finalized as Eq \ref{eq:8}:
\begin{eqnarray}
\label{eq:8}
    \textbf{h}_{i}^{l+1} = \sigma(\gamma \frac{\overrightarrow{\textbf{1}}}{N} + (1-\gamma) (\textbf{W}^l_0 \textbf{h}^l_i +\alpha \textbf{h}^l_{i,\mathrm{in}} -\beta \textbf{h}^l_{i,\mathrm{out}}))
\end{eqnarray}
where the activation function $\sigma$ is the PReLU and $\overrightarrow{\textbf{1}}$ is a 1-vector of size $d^{l+1}$. Then, the node embedding is normalized as $\textbf{h}^{l+1}_i = \frac{\textbf{h}^{l+1}_i}{||\textbf{h}^{l+1}_i||_2}$, where $||\textbf{h}^{l+1}_i||_2 =\sum_{j=1}^{d^{l+1}} (\textbf{h}_{i,j}^{l+1})^2$.

\subsection{Likelihood functions}
In traditional statistical analyses, node embedding corresponds to feature extraction.
The results of node embedding can be combined with various likelihood functions to perform various experimental tasks.

\noindent \textbf{For node classification}

After stacking BHGNN-RT layers, we fed the output $\textbf{H} \in \mathbb{R}^{N\times T}$ of the final layer with the activation function $\mathrm{log\_softmax}$ to calculate the category scores $\textbf{Z} \in \mathbb{R}^{N\times T}$.
The element $\textbf{Z}_{it}$ for node $i$ is defined as the log-softmax function $\textbf{Z}_{it} = \log(\frac{e^{\textbf{H}_{it}}}{\sum^T_{j=1} e^{\textbf{H}_{ij}}})$, representing the scoring value of node $i$ for node type $t$.

The objective function was configured as a log-likelihood function that measures the gap between the ground-truth data and predicted distribution. A smaller gap indicates stronger consistency between the two distributions.
The log-likelihood function is given by
\begin{eqnarray}
\label{eq:9} \mathcal{L}= - \frac{1}{N} \sum_i \sum^T_{t=1} y_{it} \textbf{Z}_{it}
\end{eqnarray}
\noindent where $y_{it}$ is the ground-truth label for node $i$ of node type $t$.

\noindent \textbf{For node clustering} \\ 
\indent The objective function for clustering follows the strategy in deep graph infomax (DGI) to maximize the mutual information (MI) between node representations and a global graph summary \cite{velickovic2019deep}. Recently, the Jensen-Shannon MI estimator has been proven to maximize the lower bound of MI \cite{hjelm2018learning, oord2018representation}, making the precise value of MI intractable. The Jensen-Shannon estimator works as a standard binary cross-entropy (BCE) loss, which should be maximized for the joint (positive examples) and the product of marginals (negative examples). \\
\indent We apply contrastive learning by generating a fake graph $\widetilde{\mathcal{G}}:= (\widetilde{\textbf{A}}, \widetilde{\textbf{X}})$, via row-wise shuffling of the adjacency matrix $\textbf{A}$ and initial feature matrix $\textbf{X}$. The real output of BHGNN-RT for graph $\mathcal{G}$ is $\textbf{H} \in \mathbb{R}^{N\times d^o}$, and the fake output is $\widetilde{\textbf{H}}$. In node clustering, $d^o$ is the dimension of the output layer in BHGNN-RT. A graph representation is obtained by averaging the neural representation in the graph, and the embedding result is optimized considering the global features of the graph. The global graph representation $\textbf{g} \in \mathbb{R}^{d^o}$ is defined as 
\begin{eqnarray}
\label{eq:10} \textbf{g} = f(\textbf{H}) = \mathrm{softmax}(\frac{1}{N} \sum^N_{i=1} \textbf{h}_i)
\end{eqnarray}
\indent As a proxy for maximizing the mutual information between node-graph pairwise representations, a discriminator function is applied to calculate the probabilistic score for the node-graph pair. The discriminator function is defined using a simple bilinear scoring equation \cite{mavromatis2020graph, oord2018representation} as follows:
\begin{eqnarray}
\label{eq:11} S(\textbf{h}_i,\textbf{g})=\sigma(\textbf{h}^T_i \textbf{M} \textbf{g})
\end{eqnarray}
\noindent where $\textbf{M} \in \mathbb{R}^{d^o \times d^o}$ is a learnable scoring matrix and $\sigma$ is a sigmoid function, limiting the scoring value in the range of (0,1). $\textbf{h}^T_i$ is the transpose of the node embedding $\textbf{h}_i$.
With respect to the set of original and fake graphs, the log-likelihood function is given by
\begin{eqnarray}
\label{eq:12} \mathcal{L} = \frac{1}{N} \sum_{i=1}^N (\log(S(\textbf{h}_i,\textbf{g})) + \log(1 - S(\widetilde{\textbf{h}}_i,\textbf{g}))).
\end{eqnarray}
The log-likelihood value estimates the mutual information by assigning higher scores to positive embeddings than negative ones. This objective function encourages the encoder to capture better information shared across all nodes.

\subsection{Parameter optimization}
\indent When stacking multiple layers, a central problem is the rapid growth of parameters in the weight matrices. Therefore, we adopt basis decomposition to regularize the relational matrices $\textbf{W}^l_r$ in each layer \cite{schlichtkrull2018modeling}. Matrix $\textbf{W}^l_r$ is decomposed as follows:
\begin{eqnarray}
\label{eq:13} \textbf{W}^l_r = \sum^B_{b=1}a^l_{rb} \textbf{V}^l_b,
\end{eqnarray}
\noindent which is calculated as a linear combination of the basis transformations $\textbf{V}^l_b \in \mathbb{R}^{d^{l+1}\times{d^l}}$. $\textbf{V}^l_b$ is shared across diverse relationships. Hence, only parameters $a^l_{rb}$ and the matrix $\textbf{V}^l_b$ should be learned during training. Other parameters to be optimized include the scoring matrix $\textbf{M}$ and the coefficients $\alpha$ and $\beta$ for incoming and outgoing messages.

\section{Experimental setup}
We evaluated the model performance on several benchmark datasets and compared it with the results of SOTA algorithms. Our methods were implemented on two standard tasks: multiclass node classification and clustering for directed heterogeneous graphs.
\subsection{Datasets and baselines}
\indent The model performance was evaluated on six public datasets, namely Cora \cite{mccallum2000automating}, Cora\_ml \cite{mccallum2000automating}, CiteSeer \cite{giles1998citeseer}, CiteSeer\_full \cite{giles1998citeseer}, Amazon\_CS \cite{mcauley2015image}, Amazon\_photo \cite{mcauley2015image}. These datasets are all directed heterogeneous graphs encoding directed subject-object relations. Cora, Cora\_ml, CiteSeer, and CiteSeer\_full are classical citation graphs, and Amazon\_CS and Amazon\_photo are segments of Amazon’s co-purchase graphs. Detailed statistics of the datasets are listed in Supplementary Table \ref{Supplementary Table 1}. 

\subsection{For entity classification}   
\indent For node classification, we compared our model with seven SOTA models. They are categorized into two main types: 1) spectral-based GNNs, such as ChebNet \cite{defferrard2016convolutional}, GCN \cite{kipf2016semi}, simplifying GCN (SGC) \cite{wu2019simplifying}, relational-GCN (R-GCN) \cite{schlichtkrull2018modeling}; 2) spatial-based GNNs, comprising GraphSAGE \cite{hamilton2017inductive} and graph attention network (GAT) \cite{velickovic2018graph}, directed GCN (Dir-GNN) \cite{rossi2023edge}. The mechanisms of these baselines are described in the Supporting Materials. \\
\indent The evaluation process differs subtly for node classification tasks among recent publications \cite{schlichtkrull2018modeling, hamilton2017inductive, velickovic2018graph}. We uniformly configured the experiments to eliminate these differences using a node-level random split in each graph into 70$\%$, 20$\%$, and 10$\%$ of the training, validation, and testing sets. We varied the number of layers from two to eight for each model and selected the best-performing model for the training and validation set (Figure \ref{Figure 6}). Each experiment was conducted for 10 runs with a maximum of $n=100$ epochs. Throughout the experiments, we configured the hidden layer with $h=64$ dimensions and the Adam optimizer with a learning rate of $l=0.01$. The weights were initialized using the Glorot initializer, as introduced in \cite{glorot2010understanding}. As described in the previous section, we utilized the NLL loss (Eq \ref{eq:9}) in all other baselines for the classification tasks. The parameter optimization was performed on the training set, and the optimal combination of hyperparameters was chosen for the validation set. The model performance was quantified in metrics, including accuracy and macro-F1 score (average over 10 runs).
\begin{figure}
    \centering
    \includegraphics[scale=0.4]{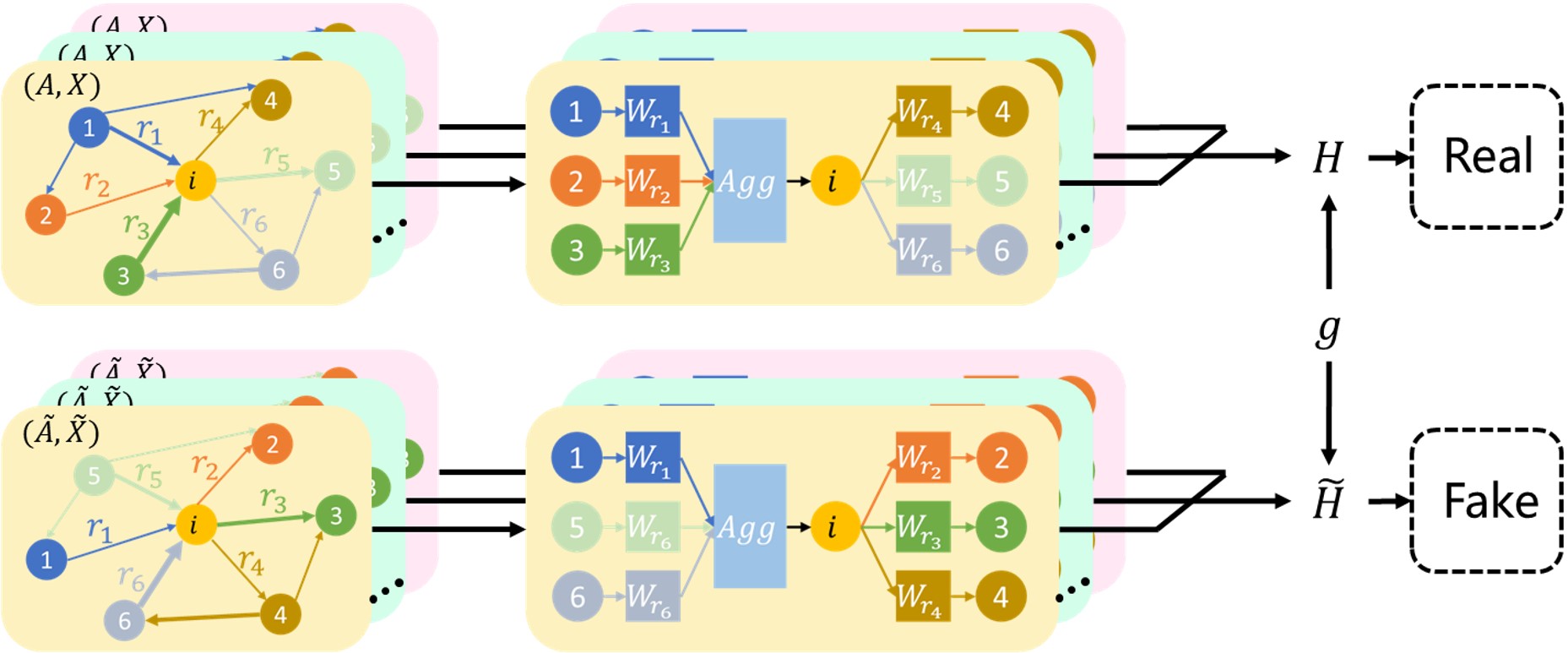}
    \caption{\textbf{BHGNN-RT framework for clustering task.} A fake graph is generated via node shuffling. The message-passing process is measured when considering input and output information flow. The objective function is to maximize the mutual information within node-graph pairwise representations.}
    \label{Figure 4}
\end{figure}

\subsection{For node clustering}  
\indent For node clustering, we utilized five benchmark methods, namely K-means, DGI \cite{velickovic2019deep}, Graph InfoClust (GIC) \cite{mavromatis2020graph}, deep attentional embedded graph clustering (DAEGC) \cite{wang2019attributed}, just balance GNN (JBGNN) \cite{bianchi2022simplifying}, and a variant of R-GCN \cite{schlichtkrull2018modeling}. Detailed descriptions of these baselines are provided in Supporting Materials. \\
\indent The general idea for clustering is to closely group nodes with similar input features in the embedding space. By stacking layers, GNNs aggregate local and global information in a graph and produce appropriate embedding results. We retained all models with 64 hidden units and 512 output units to maintain model consistency. The number of baseline layers differs, and appropriate layers were chosen based on their performance, shown as the red starred points in Figure \ref{Figure 6}. Each experiment was conducted for 10 runs with a maximum of $n=300$ epochs to obtain the results. The Adam optimizer was configured with a learning rate of $l=0.001$. \\
\indent The learned embedding results from GNNs served as the input to the K-means clustering method. Afterward, the node embeddings were grouped into $T$ clusters, and evaluations were performed by comparing the predictions and ground truth. The clustering performance was evaluated in terms of three commonly used metrics: accuracy, normalized mutual information (NMI), and adjusted Rand index (ARI) \cite{mavromatis2020graph, wang2019attributed}. NMI is a metric based on information theory and ARI is treated as an accuracy metric that penalizes incorrect predictions. \\
\indent We implemented our method and all experiments via PyTorch 1.12.0 and CUDA toolkit 11.6. The other methods were also transformed into the PyTorch platform. All experiments were conducted on a computer with a 20-core Intel i9-10900K CPU(@3.7 GHz), NVIDIA RTX A4000 GPU (16 GB of memory), and 80 GB of RAM.

\section{Results}
\indent We conducted extensive experiments for node classification and clustering tasks. The experimental results were evaluated using accuracy and macro-F1 for classification and using accuracy, NMI, and ARI for clustering. For the proposed method, further investigations analyzed the effect of the message component, model layer, and teleport proportion. In addition, we plotted the t-SNE 2D projection of the learned embedding results for the benchmark datasets and used silhouette scores (SIL) to assess model performance.  

\subsection{Node classification}
\indent All comparative results in Table \ref{Table 2} were obtained with the uniform data splitting, as introduced in Section 4. BHGNN-RT consistently outperforms all other baselines on different datasets, with a higher classification accuracy ranging from 1.8$\%$ to 11.5$\%$. The best performance of BHGNN-RT was obtained on CiteSeer\_full, with an average gain 11.5$\%$ higher than the accuracy of GAT (87.9$\pm$0.3$\%$). The results of Macro-F1 also display a similar pattern. These demonstrate the efficacy and efficiency of our proposed method. Meanwhile, when compared with BHGNN without the teleport component, the BHGNN-RT still obtains a slightly better performance. BHGNN-RT improves BHGNN's performance by at most 4.3$\%$ while on Cora, which indicates that random teleport promotes the classification capability of the proposed model.

\begin{table}[!ht]
\scriptsize
\centering
\caption{Node classification accuracy on benchmark datasets.}
\begin{tabular}{cc|ccccccc|cc}
\toprule  
               &          & ChebNet & GCN & SGC & GAT & GraphSAGE & Dir-GCN & R-GCN & BHGNN & BHGNN-RT \\
\midrule  
Cora           & Acc      & 0.767 & 0.825 & 0.889 & 0.821 & 0.845 & 0.806 & 0.923 & 0.926 & \bf0.969 \\
               &          & $\pm$0.004 & $\pm$0.002 & $\pm$0.001 & $\pm$0.008 & $\pm$0.003 & $\pm$0.003 & $\pm$0.003 & $\pm$0.021 & $\pm$\bf0.014 \\
               & Macro-F1 & 0.753 & 0.813 & 0.886 & 0.803 & 0.831 & 0.782 & 0.926 & 0.915 & \bf0.957 \\
               &          & $\pm$0.004 & $\pm$0.002 & $\pm$0.002 & $\pm$0.011 & $\pm$0.003 & $\pm$0.003 & $\pm$0.004 & $\pm$0.026 & $\pm$\bf0.017 \\
\midrule  
Cora\_ml       & Acc      & 0.831 & 0.816 & 0.822 & 0.809 & 0.836 & 0.869 & 0.906 & 0.976 & \bf0.997 \\
               &          & $\pm$0.006 & $\pm$0.003 & $\pm$0.004 & $\pm$0.005 & $\pm$0.005 & $\pm$0.003 & $\pm$0.003 & $\pm$0.011 & $\pm$\bf0.002 \\
               & Macro-F1 & 0.828 & 0.803 & 0.804 & 0.798 & 0.814 & 0.860 & 0.898 & 0.973 & \bf0.997 \\
               &          & $\pm$0.007 & $\pm$0.005 & $\pm$0.006 & $\pm$0.005 & $\pm$0.005 & $\pm$0.003 & $\pm$0.004 & $\pm$0.012 & $\pm$\bf0.002 \\
\midrule  
CiteSeer       & Acc      & 0.739 & 0.706 & 0.732 & 0.713 & 0.764 & 0.757 & 0.889 & 0.970 & \bf0.989 \\
               &          & $\pm$0.002 & $\pm$0.003 & $\pm$0.002 & $\pm$0.010 & $\pm$0.003 & $\pm$0.002 & $\pm$0.003 & $\pm$0.009 & $\pm$\bf0.003 \\
               & Macro-F1 & 0.701 & 0.671 & 0.702 & 0.684 & 0.727 & 0.724 & 0.872 & 0.965 & \bf0.987 \\
               &          & $\pm$0.003 & $\pm$0.003 & $\pm$0.002 & $\pm$0.011 & $\pm$0.004 & $\pm$0.002 & $\pm$0.004 & $\pm$0.010 & $\pm$\bf0.004 \\
\midrule  
CiteSeer\_full & Acc      & 0.803 & 0.873 & 0.850 & 0.879 & 0.834 & 0.840 & 0.857 & 0.988 & \bf0.994 \\
               &          & $\pm$0.003 & $\pm$0.002 & $\pm$0.008 & $\pm$0.003 & $\pm$0.007 & $\pm$0.004 & $\pm$0.002 & $\pm$0.002 & $\pm$\bf0.002 \\
               & Macro-F1 & 0.805 & 0.874 & 0.850 & 0.880 & 0.836 & 0.841 & 0.858 & 0.989 & \bf0.994 \\
               &          & $\pm$0.003 & $\pm$0.002 & $\pm$0.009 & $\pm$0.003 & $\pm$0.007 & $\pm$0.004 & $\pm$0.002 & $\pm$0.002 & $\pm$\bf0.002 \\
\midrule  
Amazon\_cs     & Acc      & 0.840 & 0.881 & 0.905 & 0.900 & 0.885 & 0.900 & 0.963 & 0.984 & \bf0.985 \\
               &          & $\pm$0.016 & $\pm$0.007 & $\pm$0.003 & $\pm$0.004 & $\pm$0.006 & $\pm$0.010 & $\pm$0.002 & $\pm$0.001 & $\pm$\bf0.001 \\
               & Macro-F1 & 0.769 & 0.858 & 0.886 & 0.897 & 0.860 & 0.873 & 0.960 & \bf0.984 & 0.981 \\
               &          & $\pm$0.044 & $\pm$0.011 & $\pm$0.004 & $\pm$0.005 & $\pm$0.009 & $\pm$0.018 & $\pm$0.003 & $\pm$\bf0.003 & $\pm$0.002 \\
\midrule  
Amazon\_photo  & Acc      & 0.915 & 0.943 & 0.937 & 0.940 & 0.939 & 0.942 & 0.974 & 0.985 & \bf0.992 \\
               &          & $\pm$0.017 & $\pm$0.006 & $\pm$0.002 & $\pm$0.005 & $\pm$0.008 & $\pm$0.008 & $\pm$0.002 & $\pm$0.002 & $\pm$\bf0.007 \\
               & Macro-F1 & 0.892 & 0.932 & 0.927 & 0.930 & 0.927 & 0.928 & 0.973 & 0.983 & \bf0.992 \\
               &          & $\pm$0.017 & $\pm$0.006 & $\pm$0.002 & $\pm$0.005 & $\pm$0.008 & $\pm$0.008 & $\pm$0.002 & $\pm$0.002 & $\pm$\bf0.002 \\
\bottomrule 
\end{tabular}
\label{Table 2}
\end{table}

\subsection{Node clustering}
In clustering, accuracy (ACC), NMI, and (ARI)\cite{mavromatis2020graph} were evaluated by comparing the distribution of the model predictions and ground truth. Table \ref{Table 3} summarizes the highest clustering results over 10 runs for the proposed and baseline methods. Although most recent clustering work only discussed the best results of their methods \cite{zang2021unsupervised}, we also report the results with the mean and standard deviation in the Supplementary Table \ref{Supplementary Table 3}. \\
\indent Table \ref{Table 3} demonstrates significant performance achieved by BHGNN-RT and BHGNN without teleport proportion across all datasets. BHGNN-RT exceeds its performance over other baselines. For the CiteSeer dataset, BHGNN-RT outperforms the best method, GIC, by a maximum margin of 19.3$\%$ in accuracy. For the other datasets, BHGNN or BHGNN-RT also achieves higher clustering accuracy, ranging from 4.5$\%$ to 12.2$\%$. Regarding NMI and ARI, the gain over the best benchmark methods is also significantly large across different datasets, ranging from 2.1$\%$ to 18.1$\%$ for the NMI metric and from 7.6$\%$ to 29.2$\%$ for the ARI metric. It is promising that our proposed method allows each node stronger access to the structural properties of global connectivity.

\begin{table}[!ht]
\small
\centering
\caption{Clustering performance on benchmark datasets.}
\begin{tabular}{cc|cccccc|cc}
\toprule  
  & & K-Means & DGI & DAEGC & GIC & JBGNN & R-GCN-v & BHGNN & BHGNN-RT \\
\midrule  
     & Acc & 0.408 & 0.696 & 0.547 & 0.672 & 0.483 & 0.767 & 0.876 & \bf0.889 \\
Cora & NMI & 0.215 & 0.516 & 0.371 & 0.494 & 0.384 & 0.653 & 0.758 & \bf0.790 \\
     & ARI & 0.124 & 0.470 & 0.324 & 0.400 & 0.266 & 0.559 & 0.764 & \bf0.777 \\
\midrule  
         & Acc & 0.514 & 0.699 & 0.518 & 0.647 & 0.379 & 0.507 & 0.806 & \bf0.815 \\
Cora\_ml & NMI & 0.330 & 0.512 & 0.364 & 0.485 & 0.254 & 0.467 & \bf0.660 & 0.650 \\
         & ARI & 0.226 & 0.488 & 0.285 & 0.391 & 0.167 & 0.309 & 0.625 & \bf0.633 \\
\midrule  
         & Acc & 0.440 & 0.607 & 0.602 & 0.625 & 0.468 & 0.612 & 0.791 & \bf0.818 \\
CiteSeer & NMI & 0.210 & 0.370 & 0.306 & 0.367 & 0.251 & 0.504 & 0.678 & \bf0.685 \\
         & ARI & 0.158 & 0.338 & 0.309 & 0.348 & 0.235 & 0.343 & 0.600 & \bf0.640 \\
\midrule  
               & Acc & 0.476 & 0.597 & 0.593 & 0.754 & 0.516 & 0.440 & 0.822 & \bf0.852 \\
CiteSeer\_full & NMI & 0.344 & 0.482 & 0.331 & 0.506 & 0.317 & 0.280 & 0.671 & \bf0.687 \\
               & ARI & 0.098 & 0.216 & 0.310 & 0.495 & 0.265 & 0.166 & 0.615 & \bf0.665 \\
\midrule  
           & Acc & 0.231 & 0.207 & 0.542 & 0.470 & 0.377 & 0.721 & 0.758 & \bf0.766 \\
Amazon\_cs & NMI & 0.120 & 0.029 & 0.413 & 0.461 & 0.397 & 0.772 & \bf0.793 & 0.789 \\
           & ARI & 0.064 & 0.033 & 0.395 & 0.278 & 0.215 & 0.589 & \bf0.665 & 0.658 \\
\midrule  
              & Acc & 0.275 & 0.236 & 0.651 & 0.572 & 0.513 & 0.879 & \bf0.951 & 0.944 \\
Amazon\_photo & NMI & 0.143 & 0.044 & 0.545 & 0.527 & 0.424 & 0.830 & \bf0.898 & 0.893 \\
              & ARI & 0.063 & 0.019 & 0.455 & 0.326 & 0.289 & 0.797 & \bf0.913 & 0.905 \\
\bottomrule 
\end{tabular}
\label{Table 3}
\end{table}

\subsection{Effects of message components}
\begin{figure} 
\centering
\includegraphics[scale=0.5]{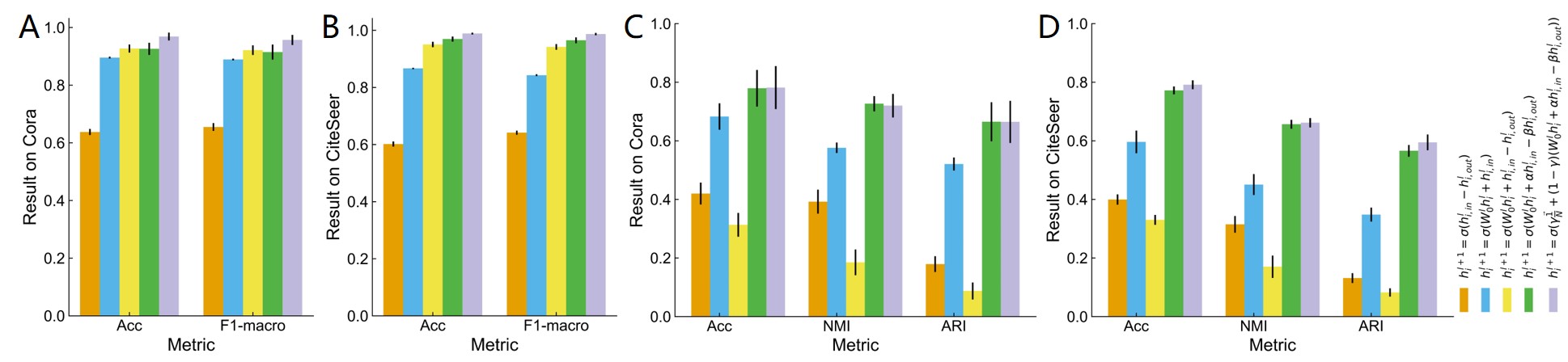}
\caption{\textbf{Evaluation of message components on model performance.} Figure 5 A-B shows the classification results, including ACC and Macro-F1, on Cora and CiteSeer, while Figure 5 C-D displays the clustering performance, including ACC, NMI, and ARI, on Cora and CiteSeer. Each bar chart exhibits results with different aggregator functions, including 1) aggregation without nodal messages, 2) aggregation without outgoing messages, 3) aggregation with unweighted incoming and outgoing messages, 4) aggregation function as Eq \ref{eq:7}, and 5) aggregation function as Eq \ref{eq:8}.} 
\label{Figure 5}
\end{figure}

\indent Previous subsections demonstrate that BHGNN-RT is an efficient encoder for directed heterogeneous graphs. Here, the intrinsic properties of BHGNN-RT are further investigated. The aggregation function in Eq \ref{eq:7} comprises the incoming, outgoing, and nodal messages. In this subsection, we discuss their functional roles in the experimental tasks. The results are presented in Figure \ref{Figure 5}.  \\
\indent In the traditional message-passing process, people mainly pay attention to the incoming messages \cite{kipf2016semi, hamilton2017inductive, velickovic2018graph}. We first tested the performance of the message-passing process via aggregation functions without nodal messages and without outgoing messages. The comparison within the first two bars, orange and blue bars, indicates that nodal messages play a more important role in model performance than outgoing messages. When integrating all message components, the classification performance gets further improved, displayed as yellow, green, and purple bars. Among them, the purple bar with the highest classification accuracy suggests that the model with aggregation function (Eq \ref{eq:8}) achieves the best performance. Differences among these three bars demonstrate the effectiveness of parameter optimization and teleport components in BHGNN-RT. \\
\indent It is notable that the major difference lies in the performance of the aggregator function with unweighted incoming and outgoing messages (yellow bars) for different experimental tasks. In the classification results (Figure \ref{Figure 5} A-B), simply integrating unweighted incoming and outgoing messages (yellow bar) can still obtain a good performance, better than orange and blue bars. However, the yellow bars perform worst, with the lowest accuracies in clustering tasks (Figure \ref{Figure 5} C-D). We assume that this phenomenon occurs because the cross-entropy loss function affects the balance between incoming and outgoing messages for node representations when without ground truth. Notably, parameters $(\alpha, \beta)$ for message components were initialized as (1, 1) for classification and (1, 0) for unsupervised clustering, to ensure faster training and better performance.    

\subsection{Effects of model layers}
\begin{figure}
\centering
\includegraphics[scale=0.48]{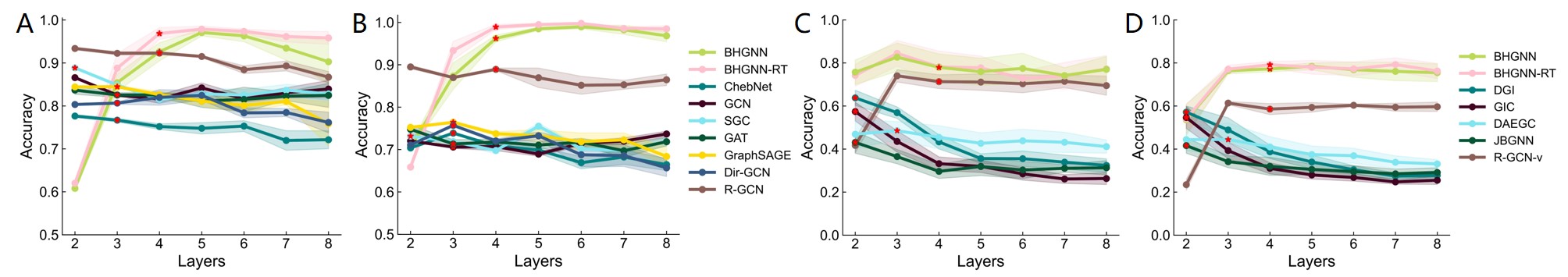}
\caption{\textbf{Effect of network layers for classification and clustering tasks.} A and B exhibit classification results, while C and D show clustering results on Cora and CiteSeer. Legends in panels B and D indicate methods used for classification and clustering tasks, respectively. The configuration of model layers is marked as the red starred points for each method.}
\label{Figure 6}
\end{figure}

Based on the message-passing framework, each node in the GNN layer can use information from its direct neighborhoods. When stacking $l$ GNN layers, each node interacts with information from the $l$-hop neighborhood \cite{barcelo2020logical}, leading to over-smoothing and overfitting \cite{xu2018representation, klicpera2019combining}. In this subsection, we evaluated the effects of the network layers on the model performance. Here, the number $l$ of network layers refers to $l-1$ convolutional layers, and we only considered the range of $l$ in [2, 8]. Except for the layer number $l$, all other model configurations were the same as those in Section 4. The experimental results are shown in Figure \ref{Figure 6}. \\
\indent In Figure \ref{Figure 6} A-B, the test accuracies increase rapidly from 2 to 4 layers for BHGNN and BHGNN-RT, demonstrating that our model performs better with more hop neighborhoods. However, other baselines exhibit a completely different trend, with the accuracies decreasing along the number of layers, which is a typical phenomenon of over-smoothing. Although the accuracy of BHGNN slightly descends after five layers, BHGNN-RT suppresses this downward trend (Figure \ref{Figure 6} A), different from the over-smoothing cases. As shown in Figure \ref{Figure 6} B, the classification results of BHGNN and BHGNN-RT are more stable after four layers when on the CiteSeer dataset. For clustering experiments, the over-smoothing phenomenon is more obvious across other benchmark methods, where the clustering accuracies decrease obviously along the number of layers.  We still observe BHGNN and BHGNN-RT display a similar pattern in Figure \ref{Figure 6} C-D and produce the best performance, with the highest accuracies. The comparison with other baselines indicates the capability of our model to overcome the over-smoothing problem to some extent.\\
\indent Regarding the hyperparameter of model layer $l$, we need to consider $l$-hop neighborhoods as a tree data structure when stacking $l$ layers. This means more layers demand larger computational complexity. In our work, to simplify the model, we configured the BHGNN and BHGNN-RT with four layers for all experiments, since their performance is relatively stable around 4 layers and higher layers do not bring a very obvious improvement. The layers of the other baselines are configured as red-starred values in Figure \ref{Figure 6}.

\subsection{Effects of teleport probability}
\begin{figure}
\centering
\includegraphics[scale=0.4]{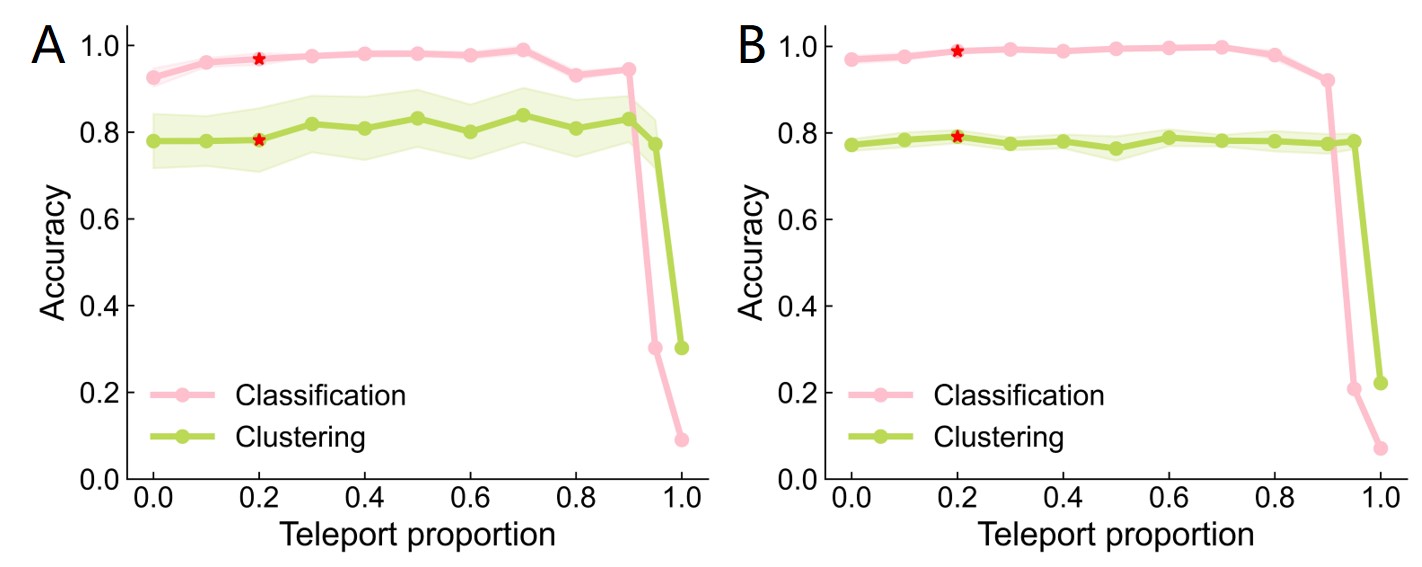}
\caption{\textbf{Effect of teleport probability on the performance of BHGNN-RT.} Panels A and B correspond to the results on Cora and CiteSeer datasets, respectively.}
\label{Figure 7}
\end{figure}

\indent In this subsection, we perturbed the teleport proportion $\gamma$ in the range of [0, 1] and investigated the BHGNN-RT performance in node classification and clustering tasks. Figure \ref{Figure 7} A and B show the results for the Cora and CiteSeer datasets. In Figure \ref{Figure 7} A, the BHGNN-RT model outperforms the BHGNN by margins of 6.4$\%$ ($\gamma=0.7$) and 6.0$\%$ ($\gamma=0.7$) for classification and clustering tasks, respectively. This highlights the effectiveness of the teleport component in BHGNN-RT. On CiteSeer dataset (Figure \ref{Figure 7} B), the highest accuracy of BHGNN-RT is obtained as 99.8$\pm$0.1$\%$ ($\gamma=0.7$) and 79.1$\pm$1.5$\%$ ($\gamma=0.2$), higher than 97.0$\pm$0.9$\%$ and 77.2$\pm$1.3$\%$ obtained via the BHGNN ($\gamma=0$) for node classification and clustering tasks. As expected from our earlier analysis, the perturbation analyses of the parameter $\gamma$ all exhibit similar tendencies across different datasets. \\
\indent While the optimal value differs slightly among different datasets, we consistently found a suitable teleport coefficient within [0.1, 0.7] to conduct experiments. The teleport proportion is adjusted according to the dataset under investigation because different graphs show diverse structures and properties. In this work, we maintained the teleport proportion at $\gamma=0.2$ to ensure the proportion of message-passing processes. The results of BHGNN-RT in Tables \ref{Table 2} and \ref{Table 3} are all with $\gamma=0.2$.

\subsection{Visualization of embedding results}
For a more intuitive comparison, we leveraged the t-SNE method \cite{van2008visualizing} to reveal the relevant embedding patterns on the Cora dataset, as shown in Figure \ref{Figure 8}. The nodes are colored based on their labels in Supplementary Table \ref{Supplementary Table 1}. Corresponding embedding results were evaluated using SIL scores, a metric to quantify the quality of the clusters generated. In Figure \ref{Figure 8}, panels A and B cannot generate obvious clustering boundaries, while panels C-F can, but their nodes belonging to different classes are still mixed in the resulting clusters. Among these eight panels, DGI, GIC, R-GCN-v, BHGNN, and BHGNN-RT share similar objective functions, like Eq \ref{eq:12}. However, only BHGNN-RT and BHGNN-RT lead to node representations that can better separate same-labeled nodes into the same group. BHGNN-RT achieves the highest SIL score of 0.477, much higher than other baseline scores. Therefore, our method improves unsupervised clustering quality when capturing more comprehensive nodal connectivity profiles and graph-level structural properties. \\
\indent In addition, the clustering performance of BHGNN-RT is evaluated across different datasets. As illustrated in Figure \ref{Figure 9}, BHGNN-RT retains a consistently favorable result among various directed heterogeneous graphs, and all nodes with various labels are separated into different clusters. Here, the Amazon datasets seem to exhibit more obvious clustering boundaries and Amazon\_photo has the highest SIL score of 0.506. We hypothesize that this phenomenon is due to the higher average degrees (around 20) of these two graphs, indicating stronger connectivity densities across their nodes.

\begin{figure}
    \centering
    \includegraphics[scale=0.5]{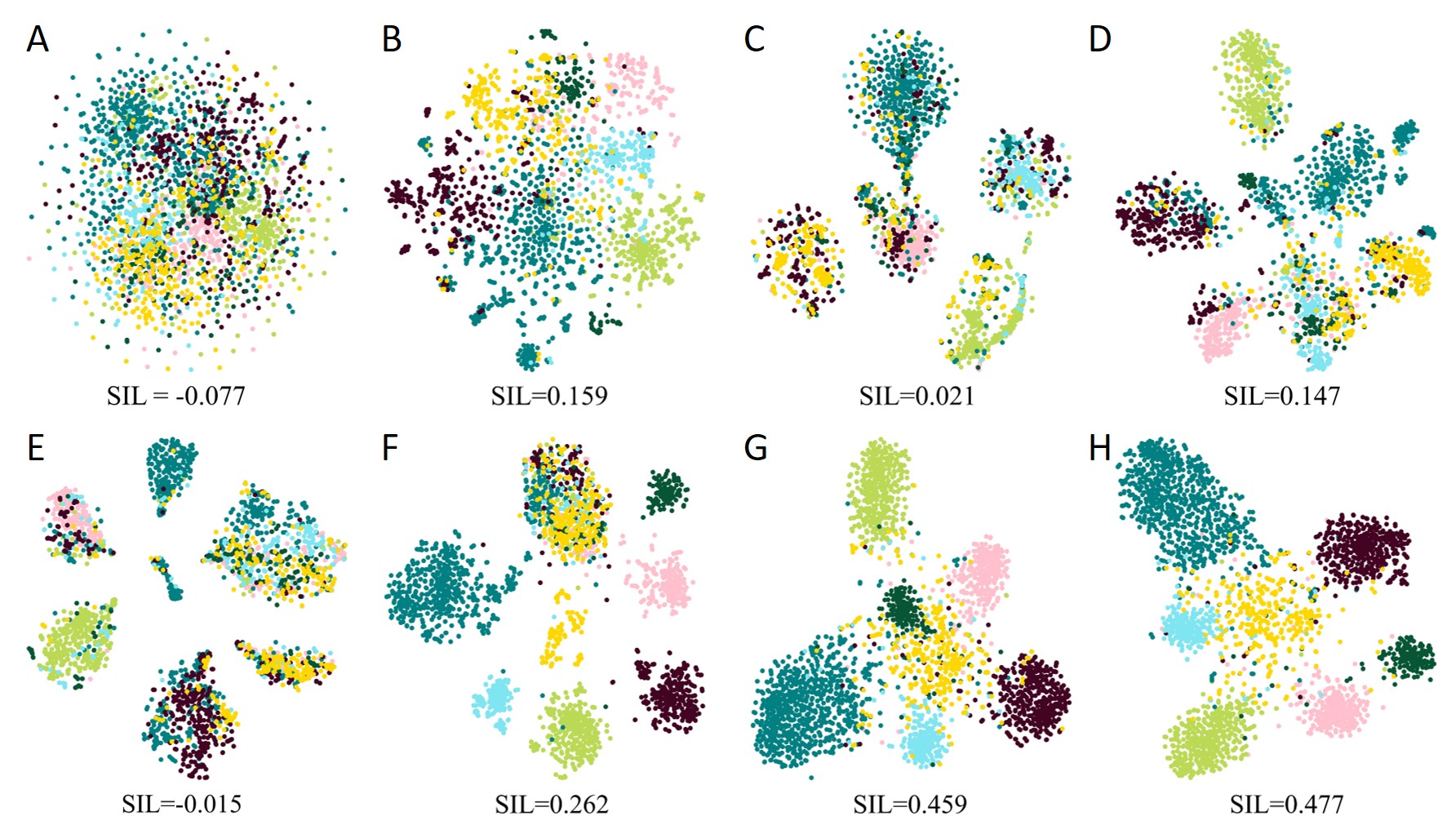}
    \caption{\textbf{t-SNE visualization for clustering Cora datasets and the corresponding SIL scores.} Individual panel depicts the results from different methods, including K-means (A), DGI (B), DAEGC (C), GIC (D), JBGNN (E), R-GCN-v (F), BHGNN (G), and BHGNN-RT (H).}
    \label{Figure 8}
\end{figure}

\begin{figure}
    \centering
    \includegraphics[scale=0.45]{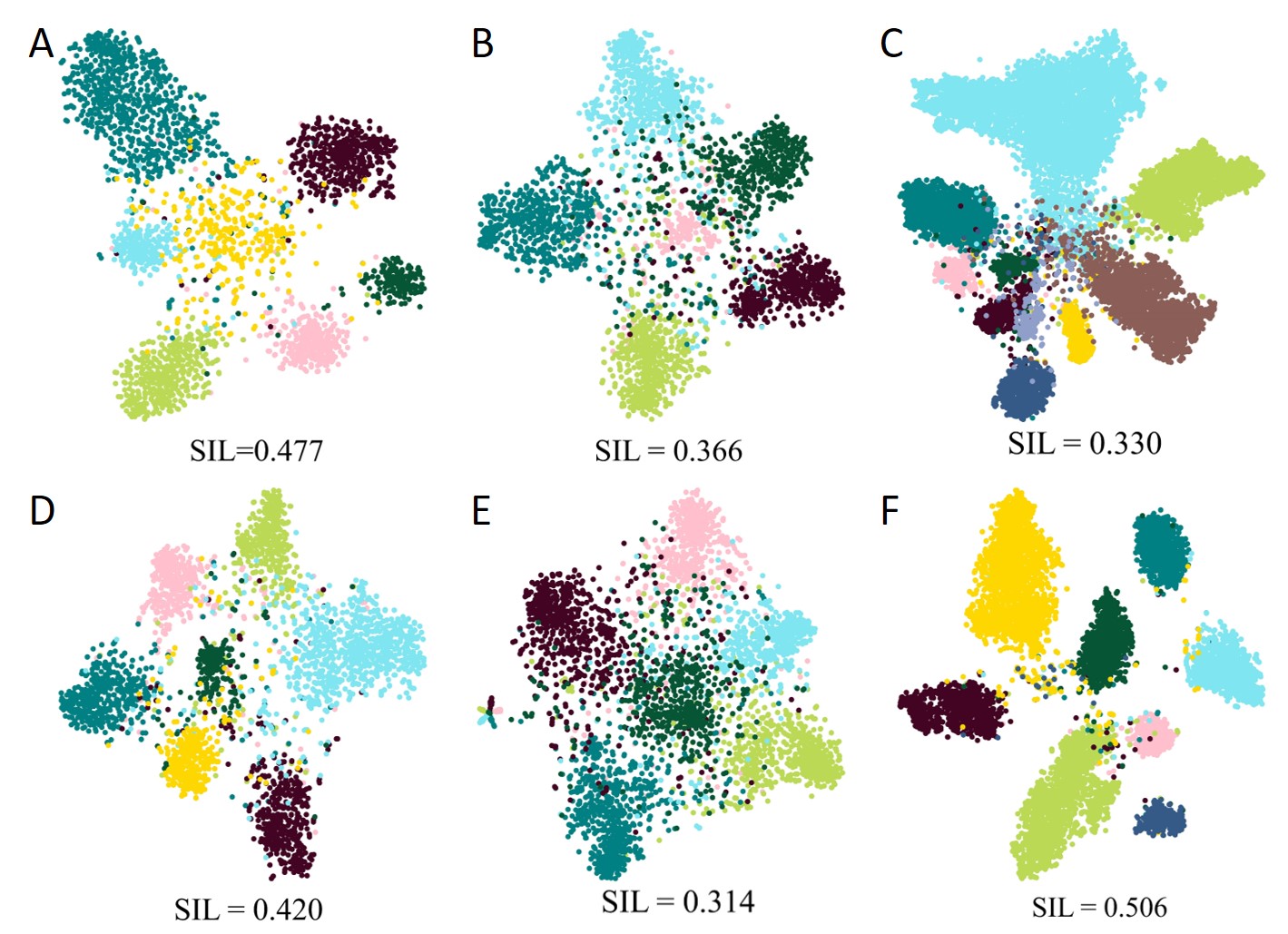}
    \caption{\textbf{t-SNE plots for clustering different datasets via BHGNN-RT and relevant SIL scores.} The datasets contain Cora (A), Cora\_ml (D), CiteSeer (B), CiteSeer\_full (E), Amazon\_cs (C), and Amazon\_photo (F).}
    \label{Figure 9}
\end{figure}

\section{Conclusion}
In this study, we first investigated the network properties of directed heterogeneous graphs, including the asymmetry between the in- and out-degree distributions and network heterogeneity. The necessity of preserving these factors was declared to ensure better graph representation learning. Accordingly, we proposed a new GNN method, named BHGNN-RT for directed heterogeneous graphs, that leverages bidirectional message-passing processes and network heterogeneity. With the optimization of teleport proportion, BHGNN-RT balances messages from existing neighborhoods with random connections, which is beneficial for overcoming the over-smoothing problem. Our method also works for unweighted and weighted graphs, allowing for more complex scenarios. \\
\indent What's more, we conducted extensive experiments to verify the efficacy and efficiency of BHGNN-RT. BHGNN-RT achieved competitive results and significantly outperformed existing baselines, both in node classification and unsupervised clustering tasks. Further investigations analyzed the effects of message components, model layer, and teleport proportion. Both nodal and outgoing messages were illustrated to promote model performance and the optimization of message coefficients is vital to ensure better performance. Introducing an appropriate teleport proportion improves the performance of BHGNN-RT and helps suppress the over-smoothing problem to some extent. Last but not least, BHGNN-RT generated more distinct clustering patterns, especially for graphs with a high average degree. In future work, we will investigate the effects of combinations of various node- and layer-wise aggregator functions and make some attempts in other more complex scenarios, such as dynamic graphs, which are not included in this work.

\section*{Acknowledgments}
This study was supported by JSPS KAKENHI Grant Number 22H00510, and AMED Grant Numbers JP23dm0207001 and JP23dm0307009.

\section*{Conflict of interest}
The authors declare no conflict of interest.

\bibliographystyle{unsrt}  
\bibliography{reference}

\clearpage
\section*{Supporting Materials}
\setcounter{table}{0}
\renewcommand{\thetable}{\arabic{table}}
\renewcommand{\tablename}{Supplementary Table}
\setcounter{figure}{0}
\renewcommand{\thefigure}{\arabic{figure}}
\renewcommand{\figurename}{Supplementary Figure}

All data generated or analyzed during this study are included in this article and its supplementary files. The code is available at https://github.com/AlbertLordsun/BHGNN-RT.
\subsection*{Benchmark datasets}
\indent Cora, Cora\_ml, CiteSeer, and CiteSeer\_full are classic citation network datasets, where nodes represent articles and edges denote citations between articles. Each article in the dataset is described by bag-of-words feature vectors, indicating the absence or presence of the corresponding words from the dictionary. Amazon\_CS and Amazon\_photo are subsets of the Amazon co-purchase network, where nodes represent goods and edges denote two kinds of goods purchased together. Similarly, the features are bag-of-words vectors extracted from product reviews. Among all the datasets, node types are given by ground-truth classes and edge types are categorized by source and target node types along the edge. The detailed statistics of the datasets are summarized in the following table. 
\begin{table}[!ht]
\centering
\caption{{Statistics of the datasets for directed heterogeneous graphs.}}
\begin{tabular}{ccccccc}
\toprule  
Datasets & Cora & Cora\_ml & CiteSeer & CiteSeer\_full & Amazon\_cs & Amazon\_photo \\
\midrule  
Nodes           &2,708    &2,995   &3,312   &4,230  &13,752   & 7,650  \\
Edges           &5,429    &8,416   &4,715   &5,358  &287,209  &143,663 \\
Classes         &7        &7       &6       &6      &10       &8       \\
Relations       &46       &49      &36      &30     &100      &64      \\
Features        &1,433    &2,879   &3703    &602    &767      &745     \\
Average Degree  &2.0      &2.8     &1.4     &1.3    &20.9     &18.8    \\ 
\bottomrule 
\end{tabular}
\label{Supplementary Table 1}
\begin{flushleft} The table lists the number of nodes, edges, node classes, edge relations, and the dimension of node features. The average degree measures the average number of edges per node in the graph.
\end{flushleft}
\end{table}

\subsection*{Distribution functions}
To better extract the distribution pattern for a degree distribution, we mainly consider its complementary cumulative distribution function (CCDF), defined as Eq \ref{eq:1}. We selected the following five common statistical distributions for fitting: We denote the lower fitting bound of the degree distribution as $x_{\mathrm{min}}$ and ensure $\int_{x_{\mathrm{min}}}^\infty C f(x) \mathrm{d}x=1$, which generates the normalization constant $C$ \cite{clauset2009power}. In our network analysis, the default value of $x_{\mathrm{min}}$ was set to 1.
\begin{table}[!ht]
\centering
\caption{Distribution functions used for fitting degree distributions.}
\begin{tabular}{c|cc}
\toprule  
 Name & $C$ & $f(x)$ \\
\midrule  
Power-law             & $(\alpha-1)x^{\alpha-1}_{\mathrm{min}}$ & $x^{-\alpha}$ \\ 
Truncated power-law   & $\frac{\lambda^{1-\alpha}}{\Gamma(1-\alpha, \lambda x_{\mathrm{min}})}$ & $x^{-\alpha} e^{-\lambda x}$ \\
Exponential           & $\lambda e^{\lambda x_{\mathrm{min}}}$ & $e^{-\lambda x}$ \\
Stretched exponential & $\beta \lambda e^{\lambda x^\beta_{\mathrm{min}}}$ & $x^{\beta-1} e^{-\lambda x^\beta}$ \\
Lognormal             & $\sqrt{\frac{2}{\pi \sigma^2}} [\mathrm{erfc}(\frac{\mathrm{ln}x_{\mathrm{min}}-\mu}{\sqrt{2}\sigma})]^{-1}$ & $\frac{1}{x} \mathrm{exp}[-\frac{(\mathrm{ln}x-\mu)^2}{2\sigma^2}]$ \\
\bottomrule 
\end{tabular}
\label{Supplementary Table 2}
\begin{flushleft} Each distribution includes the appropriate normalization constant $C$ and the basic function form $f(x)$, where $\int_{x_{\mathrm{min}}}^\infty C f(x) \mathrm{d}x=1$.
\end{flushleft}
\end{table}

\subsection*{Baselines}
\indent For node classification, the detailed descriptions of baselines are listed as follows.
\begin{itemize}
    \item \textbf{ChebNet} exploits Chebyshev polynomials to construct convolutional layers instead of time-consuming Laplacian Eigenvalue decomposition on graphs.
    \item \textbf{GCN} stacks multiple graph convolutional layers via Chebyshev polynomials and learns graph representations using a nonlinear activation function.
    \item \textbf{SGC} simplifies model complexity by successively removing nonlinearities and collapsing weight matrices between consecutive convolutional layers.
    \item \textbf{GraphSAGE} utilizes a general inductive framework that efficiently generates node embeddings by sampling and aggregating neighborhood information.
    \item \textbf{GAT} aggregates neighborhood nodal information weighted by learned attention coefficients.
    \item \textbf{Dir-GCN} extends any message-passing neural network (MPNN) to account for edge directionality information by conducting separate aggregations of the incoming and outgoing edges.
    \item \textbf{R-GCN} handles multi-relational data characteristic of realistic knowledge bases.
\end{itemize}

\indent Regarding node clustering, we include the following baselines to compare with IO-HGN-RT.
\begin{itemize}
    \item \textbf{K-means} aims to partition $n$ observations into $k$ clusters in which each observation belongs to the cluster centroid.
    \item \textbf{DGI} relies on maximizing mutual information between patch representations and corresponding high-level summaries of a graph.
    \item \textbf{GIC} is an unsupervised graph representation learning method that leverages cluster-level content.
    \item \textbf{DAEGC} encodes the topological structure and node content in an attentional manner.
    \item \textbf{JBGNN} is equipped with suitable message-passing layers and can achieve good clustering assignments by optimizing objective terms in spectral clustering.
    \item \textbf{R-GCN-v} is a model where we implant the message-passing function of the R-GCN into our architecture in Section 3.1. 
\end{itemize}

\clearpage
\subsection*{Supplementary tables}

\begin{table}[!ht]
\small
\centering
\caption{Clustering performance on benchmark datasets.}
\begin{tabular}{cc|cccccc|cc}
\toprule  
  & & K-Means & DGI & DAEGC & GIC & JBGNN & R-GCN-v & BHGNN & BHGNN-RT \\
\midrule  
               & Acc & 0.408 & 0.638 & 0.485 & 0.574 & 0.431 & 0.713 & 0.779 & \bf0.782 \\
               &     & - & $\pm$0.035 & $\pm$0.043 & $\pm$0.043 & $\pm$0.051 & $\pm$0.032 & $\pm$0.063 & $\pm$\bf0.073 \\
Cora           & NMI & 0.215 & 0.481 & 0.321 & 0.442 & 0.282 & 0.612 & \bf0.723 & 0.720 \\
               &     & - & $\pm$0.016 & $\pm$0.031 & $\pm$0.024 & $\pm$0.056 & $\pm$0.021 & $\pm$\bf0.026 & $\pm$0.040 \\
               & ARI & 0.124 & 0.398 & 0.251 & 0.333 & 0.179 & 0.524 & \bf0.665 & 0.664 \\
               &     & - & $\pm$0.036 & $\pm$0.040 & $\pm$0.035 & $\pm$0.061 & $\pm$0.019 & $\pm$\bf0.067 & $\pm$0.072 \\
\midrule  
               & Acc & 0.514 & 0.681 & 0.458 & 0.594 & 0.334 & 0.496 & \bf0.715 & 0.705 \\
               &     & - & $\pm$0.019 & $\pm$0.038 & $\pm$0.043 & $\pm$0.031 & $\pm$0.009 & $\pm$\bf0.038 & $\pm$0.042 \\
Cora\_ml       & NMI & 0.330 & 0.492 & 0.307 & 0.442 & 0.180 & 0.445 & 0.615 & \bf0.620 \\
               &     & - & $\pm$0.016 & $\pm$0.034 & $\pm$0.027 & $\pm$0.053 & $\pm$0.015 & $\pm$0.029 & $\pm$\bf0.024 \\
               & ARI & 0.226 & 0.442 & 0.220 & 0.361 & 0.108 & 0.285 & \bf0.546 & \bf0.546 \\
               &     & - & $\pm$0.038 & $\pm$0.033 & $\pm$0.034 & $\pm$0.043 & $\pm$0.015 & $\pm$\bf0.038 & $\pm$\bf0.040 \\
\midrule  
               & Acc & 0.440 & 0.571 & 0.444 & 0.547 & 0.416 & 0.586 & 0.772 & \bf0.791 \\
               &     & - & $\pm$0.028 & $\pm$0.078 & $\pm$0.051 & $\pm$0.036 & $\pm$0.026 & $\pm$0.013 & $\pm$\bf0.015 \\
CiteSeer       & NMI & 0.210 & 0.353 & 0.206 & 0.337 & 0.203 & 0.478 & 0.657 & \bf0.662 \\
               &     & - & $\pm$0.015 & $\pm$0.054 & $\pm$0.020 & $\pm$0.032 & $\pm$0.023 & $\pm$0.015 & $\pm$\bf0.016 \\
               & ARI & 0.158 & 0.274 & 0.186 & 0.296 & 0.176 & 0.321 & 0.566 & \bf0.595 \\
               &     & - & $\pm$0.043 & $\pm$0.061 & $\pm$0.033 & $\pm$0.037 & $\pm$0.016 & $\pm$0.020 & $\pm$\bf0.027 \\
\midrule  
               & Acc & 0.476 & 0.587 & 0.485 & 0.679 & 0.413 & 0.376 & \bf0.780 & 0.764 \\
               &     & - & $\pm$0.008 & $\pm$0.070 & $\pm$0.067 & $\pm$0.059 & $\pm$0.029 & $\pm$\bf0.049 & $\pm$0.071 \\
CiteSeer\_full & NMI & 0.344 & 0.468 & 0.265 & 0.473 & 0.244 & 0.237 & \bf0.582 & 0.574 \\
               &     & - & $\pm$0.006 & $\pm$0.045 & $\pm$0.025 & $\pm$0.049 & $\pm$0.038 & $\pm$\bf0.046 & $\pm$0.063 \\
               & ARI & 0.098 & 0.206 & 0.238 & 0.430 & 0.209 & 0.139 & 0.545 & \bf0.549 \\
               &     & - & $\pm$0.009 & $\pm$0.041 & $\pm$0.052 & $\pm$0.043 & $\pm$0.020 & $\pm$0.059 & $\pm$\bf0.077 \\
\midrule  
               & Acc & 0.231 & 0.206 & 0.500 & 0.423 & 0.333 & 0.694 & 0.691 & \bf0.705 \\
               &     & - & $\pm$0.001 & $\pm$0.030 & $\pm$0.026 & $\pm$0.025 & $\pm$0.030 & $\pm$0.053 & $\pm$\bf0.048 \\
Amazon\_cs     & NMI & 0.120 & 0.026 & 0.394 & 0.447 & 0.346 & 0.760 & 0.752 & \bf0.754 \\
               &     & - & $\pm$0.001 & $\pm$0.018 & $\pm$0.015 & $\pm$0.035 & $\pm$0.013 & $\pm$0.029 & $\pm$\bf0.026 \\
               & ARI & 0.064 & 0.032 & 0.331 & 0.250 & 0.186 & 0.578 & 0.594 & \bf0.606 \\
               &     & - & $\pm$0.001 & $\pm$0.044 & $\pm$0.013 & $\pm$0.022 & $\pm$0.016 & $\pm$0.067 & $\pm$\bf0.048 \\
\midrule  
               & Acc & 0.275 & 0.231 & 0.565 & 0.529 & 0.443 & 0.851 & 0.902 & \bf0.905 \\
               &     & - & $\pm$0.005 & $\pm$0.057 & $\pm$0.024 & $\pm$0.036 & $\pm$0.019 & $\pm$0.035 & $\pm$\bf0.041 \\
Amazon\_photo  & NMI & 0.144 & 0.042 & 0.481 & 0.492 & 0.379 & 0.825 & \bf0.870 & 0.869 \\
               &     & - & $\pm$0.002 & $\pm$0.046 & $\pm$0.019 & $\pm$0.020 & $\pm$0.004 & $\pm$\bf0.018 & $\pm$0.018 \\
               & ARI & 0.063 & 0.017 & 0.389 & 0.282 & 0.252 & 0.783 & \bf0.870 & \bf0.870 \\
               &     & - & $\pm$0.001 & $\pm$0.043 & $\pm$0.026 & $\pm$0.022 & $\pm$0.009 & $\pm$\bf0.034 & $\pm$\bf0.038 \\
\bottomrule 
\end{tabular}
\label{Supplementary Table 3}
\begin{flushleft} This table records the average results and standard deviations for clustering performance on 10 runs. We configured the random\_state in K-means as 0, in which case its results are the same across different runs. The best results are depicted in bold.
\end{flushleft}
\end{table}

\subsection*{Supplementary figures}
\setcounter{figure}{0}
\renewcommand{\thefigure}{\arabic{figure}}
\renewcommand{\figurename}{Supplementary Figure}

\begin{figure}
\centering
\includegraphics[scale=0.5]{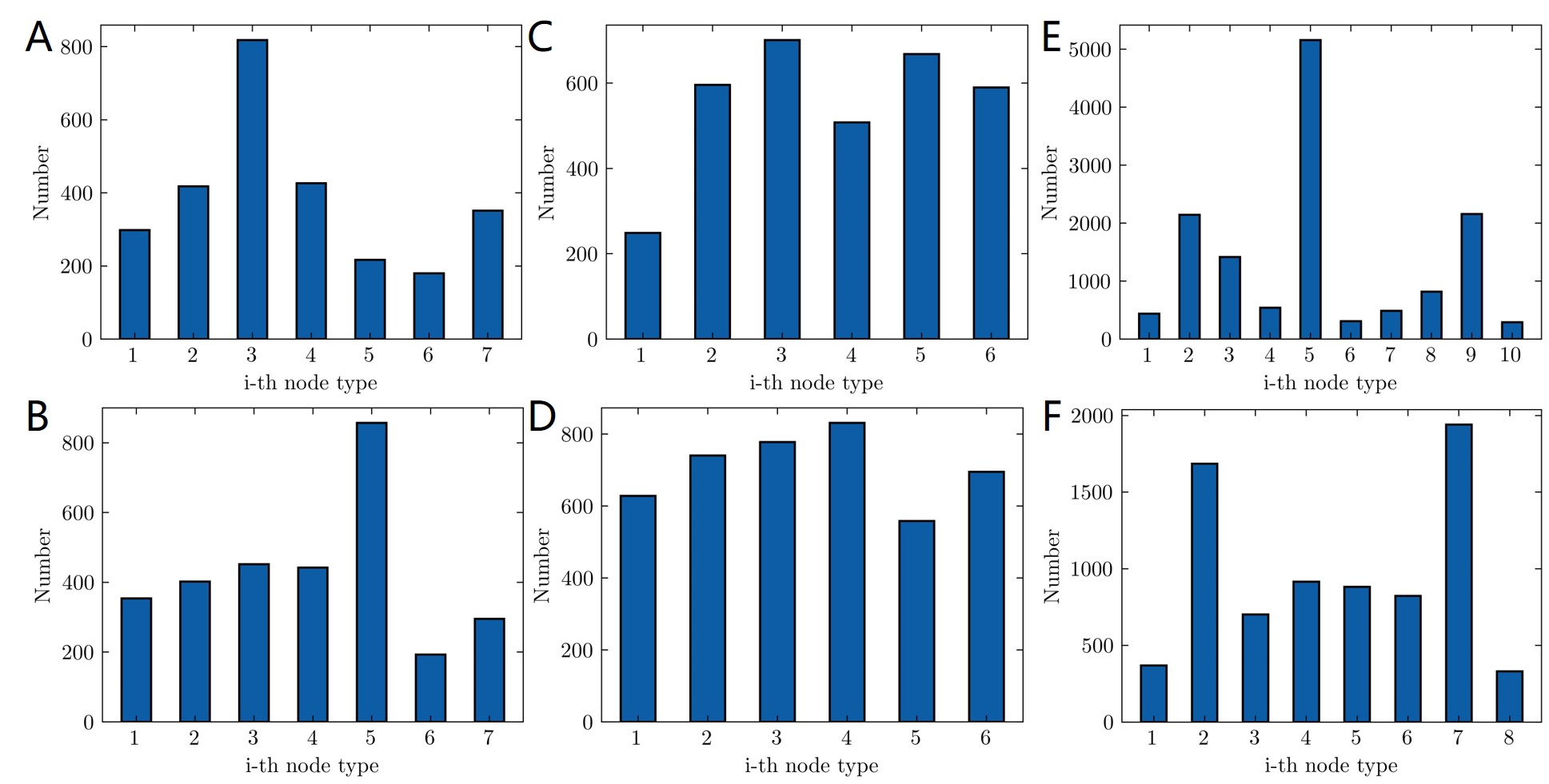}
\caption{\textbf{Heterogeneity of node types.} The datasets correspond to Cora (A), Cora\_ml (D), CiteSeer (B), CiteSeer\_full (E), Amazon\_cs (C), and Amazon\_photo (F). The horizontal axis represents the $i$-th node type in the graph.}
\label{Supplementary Figure 1}
\end{figure}

\begin{figure}
\centering
\includegraphics[scale=0.5]{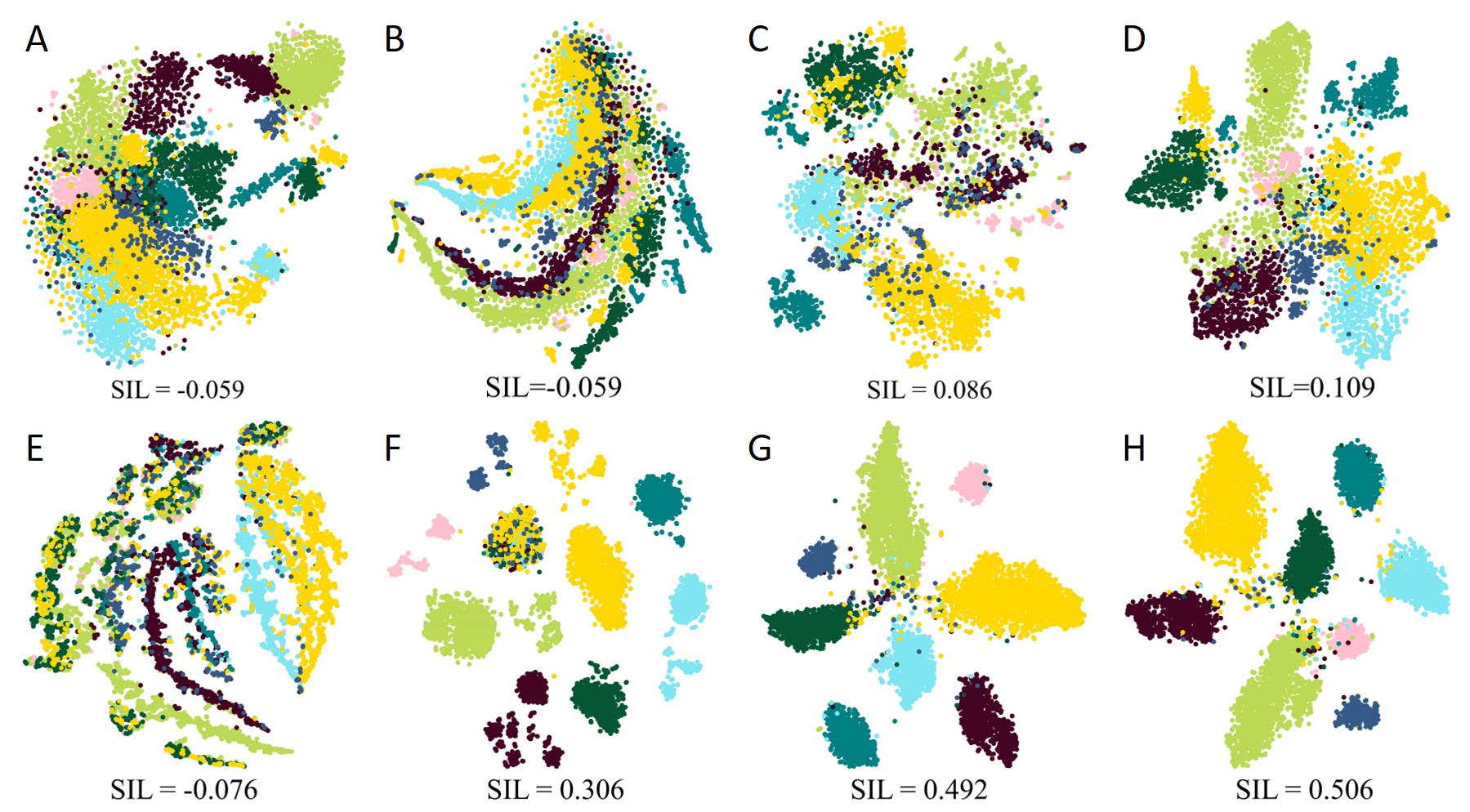}
\caption{\textbf{t-SNE visualization for clustering Amazon\_photo datasets. and the corresponding SIL scores.} Individual panel depicts the results from different methods, including K-Means (A), DGI (B), DAEGC (C), GIC (D), JBGNN (E), R-GCN-v (F), BHGNN (G), and BHGNN-RT (H).}
\label{Supplementary Figure 2}
\end{figure}

\end{document}